# Classification and Clustering of Sentence-Level Embeddings of Scientific Articles generated by Contrastive Learning

Gustavo Bartz Guedes[1,2] and Ana Estela Antunes da Silva[2]

[1]Faculdade de Tecnologia, Universidade Estadual de Campinas (UNICAMP), Limeira, Brazil
[2]Hortolândia Campus, Federal Institute of São Paulo, Hortolândia, São Paulo, Brazil

## ABSTRACT

*Scientific articles are long text documents organized into sections, each describing aspects of the research. Analyzing scientific production has become progressively challenging due to the increase in the number of available articles. Within this scenario, our approach consisted of fine-tuning transformer language models to generate sentence-level embeddings from scientific articles, considering the following labels: background, objective, methods, results, and conclusion. We trained our models on three datasets with contrastive learning. Two datasets are from the article's abstracts in the computer science and medical domains. Also, we introduce PMC-Sents-FULL, a novel dataset of sentences extracted from the full texts of medical articles. We compare the fine-tuned and baseline models in clustering and classification tasks to evaluate our approach. On average, clustering agreement measures values were five times higher. For the classification measures, in the best-case scenario, we had an average improvement in F1-micro of 30.73%. Results show that fine-tuning sentence transformers with contrastive learning and using the generated embeddings in downstream tasks is a feasible approach to sentence classification in scientific articles. Our experiment codes are available on GitHub.*

## 1. INTRODUCTION

In the scientific community, knowledge dissemination is done mainly by article publications. A scientific article is a long text document, structured in sections. Although articles may differ in their structure, they usually share a common structure, which can be described as follows.

- Background: describes the problem and the motivations for conducting the research.
- Objective: contains the purpose of the research.
- Methods: details all steps conducted in making the research to guarantee reproducibility.
- Results: reports the findings.

Although these elements might be present within articles, the corresponding section may not be explicitly defined. As a result, it becomes essential to detect and locate them within the text of an article.

In Natural Language Processing (NLP), an embedding represents a text as a fixed-size vector of real numbers [1]. An embedding can represent a single unit, such as a word, or a longer sequence, such as a sentence or an entire document. Embeddings can be learned by training a deep artificial neural network called "sentence transformer (ST) model" on a textual dataset.

Differently from a classification model that predicts a label, an ST model outputs an embedding. In this work, we fine-tuned ST models with contrastive learning to generate sentence-level embeddings from scientific articles according to their labels, as above mentioned.

A fine-tuning process involves training an existing pre-trained neural network with a particular training objective. In our work, we applied contrastive learning, which rearranges the spatial distribution of sentences in the embedding space. Thus, by using contrastive learning, sentences from the same label get closer in the embedding space, while those from different labels are pushed apart. After fine-tuning the ST model, we evaluated the generated embeddings in downstream supervised and unsupervised tasks: clustering and classification.

We used three datasets for fine-tuning, one from the computer science domain and two from the medical. From the latter, we introduce PMC-Sents-FULL, a novel dataset of sentences from the full text of scientific articles in PubMed [2].

Finally, our research investigates the use of the generated sentence embeddings from the contrastive fine-tuned models in supervised and unsupervised tasks. Therefore, our work has the following research questions (RQ):

RQ1: Does the use of fine-tuned sentence embeddings for clustering outperform using non-fine-tuned embeddings?
RQ2: The performance of traditional classifiers improves when utilizing fine-tuned sentence embeddings as opposed to non-fine-tuned embeddings?
RQ3: Does the use of fine-tuned sentence embeddings as input features for classifiers outperform a deep learning model fine-tuned directly as a sentence classifier?
Our models are available at HuggingFace and our experiments code at GitHub[1].

## 2. RELATED WORKS

In this work, we trained ST models with contrasting learning to generate sentence-level embeddings from scientific articles considering the labels: background, objective, methods, results, and conclusion.

In this section, we present the related works in three distinct groups. The first is related to sentence classification, where a classifier is trained to label each sentence. The second deals with text clustering. Finally, the last group addresses the representation of scientific articles, where a model outputs a document-level embedding.

The work of Dernoncourt and Lee [3] focuses on the Sequential Sentence Classification (SSC) task. This involves labeling each sentence given an input composed of n sequential sentences.

---
[1] https://github.com/myblindcode/sentence_embeddings

Their work introduced PubMed-RCT 20k and PubMed-RCT 200k datasets, along with the use of artificial neural networks, to SCC task.

Jin and Szolovits [4] address SCC by implementing a Hierarchical Sequential Labeling Network (HSLN) model. Their HSLN architecture has a context-enriching layer that captures and incorporates contextual information from the surrounding sentences of the input sequence. Training and testing were performed on PubMed-RCT datasets.

Cohan et al [5] present a model architecture for SCC in scientific articles abstracts. The model has a classification layer on top of each [SEP], which is a special sentence separator token in the SciBERT model [6]. Therefore, given a set of [SEP] separated sentences, each one is classified under a specific label. They also released the CSAbstruct dataset, a manually labeled sentence dataset of computer science abstracts.

Brack et al [7] present a unified approach for SCC in cross-domain scientific articles. The proposed network architecture is called SciBERT-HSLN, where they used SciBERT to generate word embeddings and the HSLN for the sentence classification. In their experiments, they used four datasets: two from abstracts and two from full articles texts. Also, a fifth was created by mixing these four datasets.

Next, we present related works that focus on text clustering, which is the task of forming groups of similar texts.

The work of Subakti et al [8] presents a comparison of text clustering between Term Frequency Inverse Document Frequency (TF-IDF) and BERT representations. TF-IDF generates a vector representation of text using only term frequency, while BERT captures the context by considering the position of terms in a text. BERT outperforms TF-IDF in 28 of 36 metrics.
Ravi and Kulkarni [9] explore four word-embedding representations using the k-means clustering algorithm on social media data in two topics. The study shows that the BERT model achieved the highest accuracy of 98% and performed best in clustering when applying the K-means algorithm.
In the subsequent paragraphs, we discuss related works belonging to the second group, which mainly centers around creating representations for scientific articles.

SPECTER [10] is a neural language model that generates document-level embeddings from the concatenation of titles and abstract texts of scientific articles. The model was created by fine-tuning SciBERT with Triplet Margin Loss, where the citation was used as a signal for the relatedness of articles. In this way, each sample consists of an anchor article, a cited article that is considered a positive sample, and a non-cited as a negative one. Thus, generated embeddings aim to capture similarity among articles.

Ostendorff et al [11] explore different sampling strategies to train models with the same similarity objective as SPECTER. Their sampling method uses the nearest neighbor strategy from a citation graph embedding and improved most of the results compared to SPECTER.
As presented, related works are based on SSC or document-level embedding generation. Most of the approaches use the texts of titles and abstracts only, and when using full texts, the datasets have a small number of articles, as in [7], with only 225 articles.
Differently, the innovations of this work compared to the presented related works are:

1. contrastive learning fine-tuning to generated sentence-level embeddings;
2. the use of these fine-tuned embeddings in downstream clustering and classification tasks;
3. release of a novel dataset with sentences extracted from the full text of 1,569 scientific articles.

## 3. METHODOLOGY

This work consisted of using the corresponding labels of the sentences present in CSAbstruct [5], PubMed-RCT 20k [3], and PMC-Sents-FULL datasets, to fine-tune ST models with contrastive learning. Subsequently, we assessed the effectiveness of using these embeddings as input features for classifier models and a clustering algorithm.

Our approach, outlined in Figure 1, consists of two stages. In the first stage, highlighted in green, we fine-tune an ST model with contrastive learning to generate sentence-level embeddings by using each sentence's corresponding label. Subsequently, we evaluate each fine-tuned ST model by encoding the sentences into embeddings and training a classifier according to the corresponding labels. As for clustering, we use the generated embeddings.

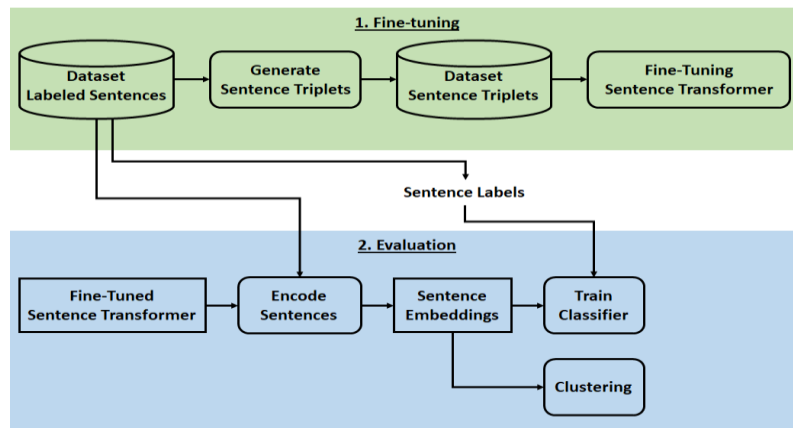

Figure 1. Classification Methodology

The next subsection 3.1 details the contrastive learning fine tuning. Subsection 3.2 presents the PMC-Sents-FULL creation. Subsection 3.3 contains the characteristics and analysis of all datasets. Finally, subsection 3.4 describes the evaluation process.

### 3.1. Contrastive learning

Contrastive learning adjusts the network weights so that similar samples get closer in the embedding space, whereas dissimilar ones get far. There are two preeminent loss functions for contrastive learning: Contrastive Loss [12] and Triplet Loss [13].

Contrastive Loss expects samples in $(a, b, \rho)$ format, where $a$ and $b$ are pairs of samples and $\rho = \{0, 1\}$ is the label, $0$ for dissimilar pairs and $1$ for similar. Triplet Loss expects samples in $(a, p, n)$ triplet format, where $a$ is an anchor sample, $p$ a similar sample in respect to $a$ and $n$ a dissimilar sample in respect to $a$.

In comparison, Contrastive Loss optimization has the downside effect of pushing all similar samples to the same point in the embedding space, because it forces the distance between similar pairs to zero. On the other hand, Triplet Loss minimizes the distance between anchors and positives while maximizing the distance between negatives, thus preserving intra-label variance and better adjusting in the presence of outliers.

The Triplet Loss function is presented in Equation 1, where $d$ is a distance function (e.g. euclidian), $a$ the anchor, $p$ a similar positive sample, $n$ a dissimilar negative sample and $m$ is an enforcing margin between $p$ and $n$.

$$\mathcal{L}_{tri} = \max(d(a,p) - d(a,n) + m, 0) \quad (1)$$

In this work, the fine-tuning approach was based on [14] which used Triplet Loss in a Bi-directional Long Short-Term Memory (Bi-LSTM) neural net architecture in Wikipedia articles. They stated, "that sentences belonging to the same section are typically more thematically related than sentences appearing in different sections".

Sentence BERT (SBERT), from [15], is a framework that generates sentence and short text-level embeddings. Among the available losses, *BatchAllTripletLoss* accepts batches of samples in the (s, ρ) format, where s is a sentence and ρ the corresponding label. We fine-tuned our models with *BatchAllTripletLoss* function.

*BatchAllTripletLoss* calculates the loss for every valid triplet present in a batch. This means that it treats ρ of the same value as similar samples and ρ with different values as dissimilar. The concept behind *BatchAllTripletLoss* is that through prolonged iterations over the complete dataset, all potential positive pairs become observable.

## 3.2. PMC-Sents-FULL Dataset Creation

PMC-Sents-FULL is based on PubMed Central, which provides open access to a subset of medical articles in XML format [2]. We have downloaded about 300,000 articles and performed the following procedures to generate the final dataset:

1. Filtered articles that have section titles for the four labels (*background*, *objective*, *methods*, and *results*). We used a keyword list with exact matching and kept only articles that had all four labels;
2. Cleaned the text by replacing references of tables and figures with markers i.e. @*table* and @*fig*;
3. Used spaCy, from [16], to split section texts into sentences;
4. Labeled each sentence using the corresponding section title as the label, including a fifth others label.

Finally, due to the filter applied in step 1, PMC-Sents-FULL has a total of 1,569 articles and 173,092 sentences. PMC-Sents-FULL was stratified split in 80%/10%/10% for train/validation/test sets (138,473; 17,309; 17,310). In the next section, we present an analysis of the three datasets used in this work.

## 3.3. Datasets Characteristics and Analysis

Table 1 summarizes the characteristics of the three datasets used in fine-tuning our models: CSAbstract [5], PubMed-RCT 20k [3], and PMC-Sents-FULL.

Table 1. Datasets Characteristics.

| Dataset | Domain | #Papers | Type | Labels |
| --- | --- | --- | --- | --- |
| PubMed-RCT 20k | Biomedicine | 20,000 | Abstract | Background, Objective, Methods, Results, Conclusion |
| CSAbstruct | Computer Science | 2,189 | Abstract | Background, Objective, Methods, Results, Other |
| PMC-Sents-FULL | Medical | 1,569 | Full Text | Background, Objective, Methods, Results, Other |

When analyzing the datasets, we found samples with the same sentence text but with different labels: 85 in CSAbstruct and 161 in PubMed-RCT 20k. We excluded these samples from all training, validation, and test sets.

The label distribution of the training sets is presented in Table 2. From this, we present the following highlights:

- The three datasets share the *background*, *objective*, *methods*, and *results* labels;
- PubMed-RCT 20k is the only one that has the *conclusion* label, but it lacks the *other* label. However, an *other* label is essential to identify sentences outside of the target labels.
- PMC-Sents-FULL considers the full text of articles, thus, it has a much higher proportion of *others* labels.
- Comparing CSAbstruct and PMC-Sents-FULL, if we shift the labels for binary values of relevant (*background*, *objective*, *methods*, and *results*) and irrelevant (*others*), in CSAbstruct we have 99,40% relevant and only 0,60% of irrelevant samples whereas in PMC-Sents-FULL there is a more even distribution of 46,34% relevant and 53,66% irrelevant.

Table 2. Label Distribution in train sets.

| Dataset | Label | #Sents |
|---|---|---|
| PubMed-RCT 20k | Methods | 59,309 (33%) |
| | Results | 57,908 (32%) |
| | Conclusion | 27,134 (15%) |
| | Background | 21,709 (12%) |
| | Objective | 13,832 (8%) |
| CSAbstruct | Methods | 3,611 (32%) |
| | Background | 3,603 (32%) |
| | Results | 3,603 (32%) |
| | Objective | 1,321 (12%) |
| | Other | 66 (3%) |
| PMC-Sents-FULL | Other | 74,304 (54%) |
| | Methods | 32,762 (24%) |
| | Results | 16,314 (12%) |
| | Background | 10,265 (7%) |
| | Objective | 4,828 (3%) |

Lastly, since we used Triplet Loss for fine-tuning, as detailed in Section 3.1, we created the validation sets with one triplet per sample, randomly selecting the positive and negative samples.

### 3.4. Models Evaluation

To evaluate our models, we compared the embeddings generated from baseline (non-fine-tuned) and fine-tuned models using five classification algorithms and one clustering algorithm.

For the sentence classification, as in [5] and [7] related works, we measured and reported the F1-micro metric.

For clustering, we evaluate the fine-tuned models with Adjusted Rand Index (ARI) and Adjusted Mutual Information (AMI) clustering agreement measures, and the Silhouette Score (Sil ). ARI and AMI are agreement measures between a generated and a known partition. Thus, we used the ground truth labels of the test set as the known partition. As for the Sil, it measures the cohesion of samples within a cluster and the separation of clusters by computing the mean distances of intra and inter clusters samples.

## 4. EXPERIMENTS

In this section, we present the fine-tuning training of ST models in the three datasets presented in Section 3 and our evaluation approaches.

**Training**

We have selected two models for fine-tuning: SciBERT2 and all-MiniLM-L6-v2 (MiniLM). SciBERT, from [6], was chosen because it is a BERT-based language model for the scientific domain. It was trained on the full text of 1.14 million articles from the Semantic Scholar database [17]. It is important to note that SciBERT has an input limit of 512 tokens and generates embeddings with 768 dimensions.

MiniLM is a language model trained with knowledge distillation, a technique that compresses a larger model (the teacher) into a smaller model (the student). The student has a similar or even better performance than the teacher, but it has fewer parameters, consuming less computational resources to fine-tune and make inferences. The all-MiniLM-L6-v2 is an ST model fine-tuned from the pre-trained MiniLM model of [18] on over 1 billion pairs of samples. It has an input limit of 256 tokens and generates embeddings with 384 dimensions.

The models were fine-tuned with the following parameters:

- *Epochs*: 20.
- *Batch Size*: 32 for all-MiniLM-L6-v2 (all datasets); 32 for SciBERT (CSAbstruct only); and 16 for SciBERT in remaining datasets.
- *Learning Rate*: 2e-5.
- *Warm up*: 10% of training data.
- *Pooling Method*: mean pooling.
- *Loss*: BatchAllTripletLoss.

SciBERT and MiniLM were also fine-tuned as classifiers, that is, label prediction instead of embedding generation, in all three datasets. These were trained for 20 epochs with a batch size of 16, and the [CLS] token was used as the sentence representation for classification.

### 4.1. Evaluation

Our approach for evaluation consisted in comparing the sentence embeddings generated with baseline and fine-tuned ST models using clustering and classification.

To perform clustering, we utilized the k-means algorithm on the test sets and calculated ARI, AMI, and Silhouette scores. ARI and AMI were computed using the partition generated by the k-means algorithm and the known partition was the labelled test set. We set the total number of clusters to be equal to the total of distinct labels present in each corresponding test set.

For the classification task, we use the generated sentence embeddings of ST fine-tuned models as the input features to train each classifier. As for the training parameters of the classifiers, only the default parameters of KNN were changed. The square root of total samples was used as the *number of neighbors*, and the *weight function* was set to "distance", this way closer neighbors have more influence when assigning the label.

Finally, F1-micro was computed using each trained classifier and the results are presented in the next section.

## 5. RESULTS

Clustering and classification results are presented in Tables 3 and 4, respectively. The bold marks the highest values within each ST model (MiniLM or SciBERT) and the underline the highest values among all classifiers, regardless of the ST model.

The first research question (*RQ1*) assesses the impact of fine-tuning in clustering. Thus, Table 3 presents the ARI, AMI, and Sil values measured in each test set. Table 3. Clustering agreement measures in the test sets.

| Dataset → | CSAbstruct | | | PubMed-RCT | | | PMC-Sents-FULL | | |
|---|---|---|---|---|---|---|---|---|---|
| Metric→ | ARI | AMI | Sil | ARI | AMI | Sil | ARI | AMI | Sil |
| *MiniLM* | | | | | | | | | |
| Baseline | 01.11 | 2.01 | 2.81 | 6.40 | 8.49 | 1.95 | 0.21 | 4.90 | 2.57 |
| Fine-Tuned | **28.69** | **31.91** | **84.39** | **73.49** | **66.78** | **98.98** | **36.36** | **30.17** | **97.51** |
| *SciBERT* | | | | | | | | | |
| Baseline | 6.77 | 13.05 | 3.16 | 22.12 | 32.67 | 5.42 | 5.34 | 10.99 | 5.40 |
| Fine-Tuned | **<u>50.72</u>** | **<u>47.32</u>** | **30.45** | **<u>76.09</u>** | **<u>67.62</u>** | 51.60 | **<u>40.62</u>** | **<u>33.94</u>** | 27.28 |

To provide visual information, we used t-distributed Stochastic Neighbor Embedding (t-SNE) to plot the samples of CSAbstruct and PubMed-RCT in a 2D graph. Figure 2 gives an overall picture of sample distribution in a 2D embedding space for the baseline and fine-tuned SciBERT models. This visualization shows a considerable improvement in clustering.

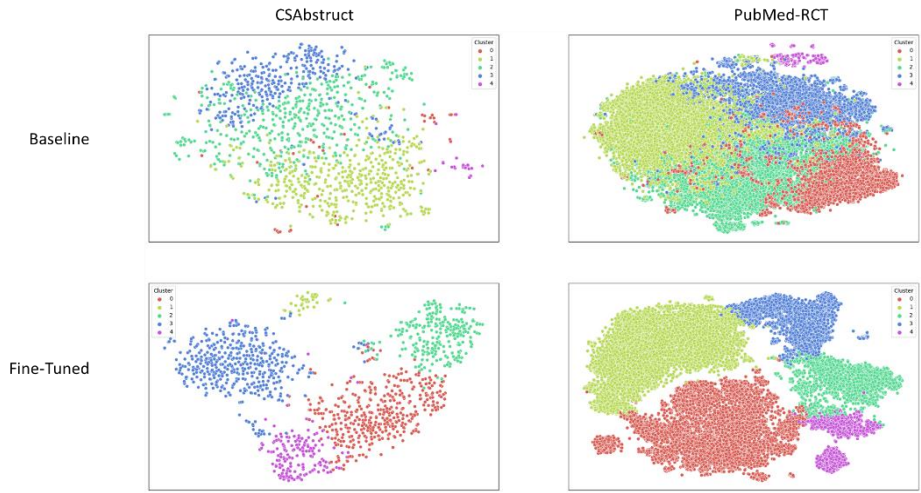

Figure 2. Clustering results with SciBERT

Research questions *RQ2* and *RQ3* are related to the performance of using the embeddings of the fine-tuned models for classification. We present classifier results for each test set using F1-micro in Table 4. *RQ3* also required fine-tuning SciBERT and MiniLM as classifier models. We use the [CLS] token as the sentence representation. In Table 4 these are rows 6 and 17.

In the next Section 6, we discuss these results and in Section 7 we present our conclusions and future works.

## 6. DISCUSSION

In this section we discuss the results in relation to the research questions *RQ1*, *RQ2*, and *RQ3*.
First, we see a huge improvement in all clustering metrics of fine-tuned models, as shown in Table 3. Considering SciBERT, on average, the values are more than five times higher than baseline models. Comparing SciBERT to MiniLM, we observe that SciBERT has higher ARI and AMI, but a lower Sil. Therefore, fine-tuning with Triplet Loss greatly improves clustering, which answers *RQ1*.

For classification, when comparing results from the same classifiers, the fine-tuned embeddings outperform all but one baseline model, which is SVM for CSAbstruct. SciBERT baseline had superior performance, 76.20 against 75.76 (rows 16 and 22 respectively of Table 4). However, the difference is only 0.44.

Table 4. F1-micro in test sets.

| ST Model ↓ | Dataset → | CSAbstruct | PubMed-RCT | PMC-Sents-FULL |
|---|---|---|---|---|
| *MiniLM* | | | | |
| **Baseline** | **Classifier ↓** | | | |
| (1) | Decision Tree | 41.59 | 44.54 | 46.21 |
| (2) | KNN | 57.89 | 61.61 | 62.72 |
| (3) | MLP | 57.89 | 74.24 | 64.01 |
| (4) | Random Forest | 55.37 | 62.01 | 60.32 |
| (5) | SVM | 61.23 | 77.99 | 69.55 |
| **Fine-Tuned** | | | | |
| (6) | MiniLM [CLS] | 68.05 | 82.40 | 70.02 |
| (7) | Decision Tree | 61.45 | 79.61 | 60.01 |
| (8) | KNN | 74.65 | 85.19 | 71.42 |
| (9) | MLP | 70.79 | 82.99 | 71.39 |
| (10) | Random Forest | **75.02** | **85.65** | **71.54** |
| (11) | SVM | 61.90 | 83.00 | 71.37 |
| *SciBERT* | | | | |
| **Baseline** | Classifier ↓ | | | |
| (12) | Decision Tree | 50.19 | 64.54 | 52.94 |
| (13) | KNN | 71.68 | 82.60 | 66.76 |
| (14) | MLP | 68.57 | 81.73 | 66.69 |
| (15) | Random Forest | 69.61 | 81.74 | 66.49 |
| (16) | SVM | 76.20 | 86.75 | 72.01 |
| **Fine-Tuned** | | | | |
| (17) | MiniLM [CLS] | 66.42 | 82.12 | 69.97 |
| (18) | Decision Tree | 68.57 | 82.28 | 71.31 |
| (19) | KNN | 77.61 | 87.05 | 75.76 |
| (20) | MLP | 70.72 | 86.29 | 73.25 |
| (21) | Random Forest | **78.50** | **87.19** | 75.89 |
| (22) | SVM | 75.76 | 87.12 | **76.27** |

In tables 5, 6, and 7 we present comparison data for classification in tabular format. There is a delta percentage (Δ%) column in all tables that displays the difference between the fine-tuned and baseline models.

In Table 5, the results for the best classifiers are presented for each dataset. As a result of fine-tuning, both MiniLM and SciBERT have improved. Notably, MiniLM exhibited an average improvement of over three times that of SciBERT, with percentages of 30.73% and 8.45%, respectively. This discrepancy can be attributed to the fact that SciBERT was trained in scientific texts and has a specific vocabulary, so it is already more adapted to this domain. This observation underscores the significant impact that vocabulary alone can wield over embedding representations in the context of classification.

| ST Model / Dataset | Best CLS Model | F1-Micro | | Δ% |
|---|---|---|---|---|
| | | Fine-Tuned | Baseline | |
| *MiniLM* | | | | |
| CSAbstruct | Random Forest | 75.02 | 55.37 | 35.48% |
| PubMed-RCT | Random Forest | 85.65 | 62.01 | 38.11% |
| PMC-Sents-FULL | Random Forest | 71.54 | 60.32 | 18.60% |
| | | | Average | **30.73%** |
| | | | Std. | 8.64 |
| *SciBERT* | | | | |
| CSAbstruct | Random Forest | 78.50 | 69.61 | 12.78% |
| PubMed-RCT | Random Forest | 87.19 | 81.74 | 6.66% |
| PMC-Sents-FULL | SVM | 76.27 | 72.01 | 5.92% |
| | | | Average | **8.45%** |
| | | | Std. | 3.07 |

Table 5. Best classifiers models comparison.

Contrastive learning fine-tuning greatly benefits classifiers that use distance as the similarity measure for inference. This is the case with KNN, which, at inference time, searches for the closest similar sample in the vector space. Therefore, Table 6 compares the KNN performance of fine-tuned and baseline models. It shows that KNN has had a considerable improvement. We also introduced a last column that shows small differences between KNN and the best classifiers in Table 5.

With the results of Tables 5 and 6, we conclude that fine-tuned embeddings for classification outperform the performance of non-fine-tuned embeddings, answering *RQ2*.

Table 6. KNN classifiers comparison.

| ST Model / Dataset | F1-Micro Fine-Tuned | F1-Micro Baseline | Δ% | Δ% Best Model |
|---|---|---|---|---|
| *MiniLM* | | | | |
| CSAbstruct | 74.65 | 57.89 | 22.44% | -0.50% |
| PubMed-RCT | 85.19 | 61.61 | 27.68% | -0.54% |
| PMC-Sents-FULL | 71.42 | 62.72 | 12.17% | -0.17% |
| | | Average | 20.76% | -0.40% |
| | | Std. | 6.44 | 0.17 |
| *SciBERT* | | | | |
| CSAbstruct | 77.61 | 71.68 | 7.64% | -1.15% |
| PubMed-RCT | 87.05 | 82.60 | 5.11% | -0.16% |
| PMC-Sents-FULL | 75.76 | 66.76 | 11.88% | -0.68% |
| | | Average | 8.21% | –0.66% |
| | | Std. | 2.79 | 0.40 |

Finally, Table 7 is a comparison between our approach and SciBERT and MiniLM language models trained as sentence classifiers. Our approach outperforms these classifiers in all datasets. Which answers *RQ3* positively.

Table 7. Comparison best models with [CLS] Fine-Tuned.

| ST Model / Dataset | Best Classifier | F1-Micro ST Fine-Tuned | F1-Micro LM Fine-Tuned | Δ% |
|---|---|---|---|---|
| *MiniLM* | | | | |
| CSAbstruct | Random Forest | 75.02 | 68.05 | 9.29% |
| PubMed-RCT | Random Forest | 85.65 | 82.40 | 3.79% |
| PMC-Sents-FULL | Random Forest | 71.54 | 70.02 | 2.12% |
| | | | Average | 5.07% |
| | | | Std. | 3.06 |
| *SciBERT* | | | | |
| CSAbstruct | Random Forest | 78.50 | 66.42 | 15.39% |
| PubMed-RCT | Random Forest | 87.19 | 82.12 | 5.81% |
| PMC-Sents-FULL | SVM | 76.27 | 69.97 | 8.26% |
| | | | Average | 9.82% |
| | | | Std. | 4.06 |

## 7. CONCLUSION

This work presented an approach to sentence classification of scientific articles by fine-tuning ST models with contrastive learning and using the generated embeddings for clustering and classification. The fine-tuned models were evaluated against two baseline (non-fine-tuned) models, specifically SciBERT and MiniLM.

As for MiniLM and SciBERT, the latter has the overall best performance. We point to the following reasons: a) SciBERT has a vocabulary from the scientific domain; b) SciBERT supports an input size that is the double of MiniLM (512/256) and outputs an embedding twice the size (768/384).

In respect to the three research questions presented in the introduction (Section 1), we have a positive response to all of them, since fine-tuned models have outperformed baseline models.
Lastly, for future work, it is important to assess the results with larger batch sizes since this creates more variability in the loss function and impacts the training process. Also, define new methods to enrich the semantics of generated embeddings considering a larger context, and finally, explore other contrastive loss functions.

## ACKNOWLEDGEMENTS

This study was financed in part by the Coordenação de Aperfeiçoamento de Pessoal de Nível Superior - Brasil (CAPES) - Finance Code 001.

## AUTHORS

**Gustavo Bartz Guedes** has a master's degree on Information Technology from University of Campinas, Brazil. He is currently a PhD student at University of Campinas and professor at Federal Institute of São Paulo. His research interests are in Artificial Intelligence and NLP solutions applied to scientific knowledge discovery.

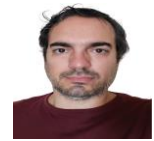

**Ana Estela Antunes da Silva** has a PhD in Computer Engineering from University of Campinas. She is currently an associate professor at University of Campinas. Her research interests are in machine learning, data mining and support decision systems.

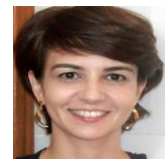